\documentclass[review]{elsarticle}

\usepackage{lineno,hyperref}
\modulolinenumbers[5]

\usepackage{color}

\usepackage{booktabs}
\usepackage{comment}
\usepackage{amsmath,amssymb}
\usepackage{diagbox}
\usepackage{multirow}
\usepackage{makecell}
\usepackage{algorithm}
\usepackage{algorithmic}
\usepackage{bbm}
\usepackage{subcaption}

\journal{Journal of \LaTeX\ Templates}

\begin{document}

\begin{frontmatter}

\title{Domain Consistency Regularization for Unsupervised Multi-source Domain Adaptive Classification}

\author[1,2]{Zhipeng Luo}
\ead{zhipeng001@e.ntu.edu.sg}
\author[3]{Xiaobing Zhang}
\ead{zxbing_uestc@163.com}
\author[1]{Shijian Lu\corref{corresponding}}
\ead{shijian.lu@ntu.edu.sg}
\cortext[corresponding]{Corresponding author}
\author[2]{Shuai Yi}
\ead{yishuai@sensetime.com}

\address[1]{Nanyang Technological University, Singapore}
\address[2]{Sensetime Research, 182 Cecil Street, 36-02 Frasers Tower, Singapore}
\address[3]{University of Electronic Science and Technology of China, China}

\begin{abstract}
Deep learning-based multi-source unsupervised domain adaptation (MUDA) has been actively studied in recent years. Compared with single-source unsupervised domain adaptation (SUDA), domain shift in MUDA exists not only between the source and target domains but also among multiple source domains. Most existing MUDA algorithms focus on extracting domain-invariant representations among all domains whereas the task-specific decision boundaries among classes are largely neglected. In this paper, we propose an end-to-end trainable network that exploits domain Consistency Regularization for unsupervised Multi-source domain Adaptive classification (CRMA). CRMA aligns not only the distributions of each pair of source and target domains but also that of all domains. For each pair of source and target domains, we employ an intra-domain consistency to regularize a pair of domain-specific classifiers to achieve \textit{intra-domain alignment}. In addition, we design an inter-domain consistency that targets joint \textit{inter-domain alignment} among all domains. To address different similarities between multiple source domains and the target domain, we design an authorization strategy that assigns different authorities to domain-specific classifiers adaptively for optimal pseudo label prediction and self-training. Extensive experiments show that CRMA tackles unsupervised domain adaptation effectively under a multi-source setup and achieves superior adaptation consistently across multiple MUDA datasets.
\end{abstract}


\end{frontmatter}


\section{Introduction}
In recent years, deep neural networks have brought great improvements to a variety of visual learning tasks, such as classification \cite{he2016deep}, segmentation \cite{chen2017rethinking, huang2021cross}, and detection \cite{ren2015faster, zhang2021meta}. These achievements mainly attribute to the availability of large-scale labeled data for supervised learning. 
However, it is prohibitively labor-intensive and time-consuming to collect abundant labeled data for each new task. 
Domain Adaptation (DA) aims to tackle this problem by utilizing labeled data in relevant domains. Specifically, it leverages a label-rich domain(s) (i.e., source domain(s)) to learn a discriminative model that generalizes well on a label-scarce domain (i.e., target domain). Most DA methods focus on single-source unsupervised domain adaptation (SUDA), where the labeled data in a single source domain are adapted via discrepancy minimization \cite{tzeng2014deep, yan2017mind}, adversarial learning \cite{ganin2015, Tzeng2017Adversarial}, prototypical networks \cite{pan2019transferrable}, etc. 

In real life, information often comes in a wide variety of formats and origins and this makes the learning process more complicated. Multi-source unsupervised domain adaptation (MUDA) aims to adapt from multiple labeled source domains of different distributions to a single target domain. It has been tackled by learning domain invariant features and predicting pseudo labels for target-domain samples and has achieved promising results in various benchmarks \cite{Xu2018Deep,Zhao2019Multi,zhao2020,wang2020learning, wang2019tmda}. On the other hand, existing methods have some common constraints. First, using domain classifiers (e.g., discriminators) to learn domain invariant features \cite{zhao2018a,wang2019tmda,zhao2020} tends to suffer from over-alignment problems since it neglects the task-specific decision boundaries of different categories. Second, predicting pseudo labels for target-domain samples \cite{Xu2018Deep} often suffers from label noises due to the different distributions of source-domain samples and the trained models. It often requires heuristic thresholds for identifying high-confidence predictions. However, selecting heuristic thresholds is a challenging task in MUDA where multiple source domains often have different similarities with the target domain and require different heuristic thresholds for optimal pseudo label prediction.

\begin{figure}[t]
  \centering
  \includegraphics[width=\textwidth]{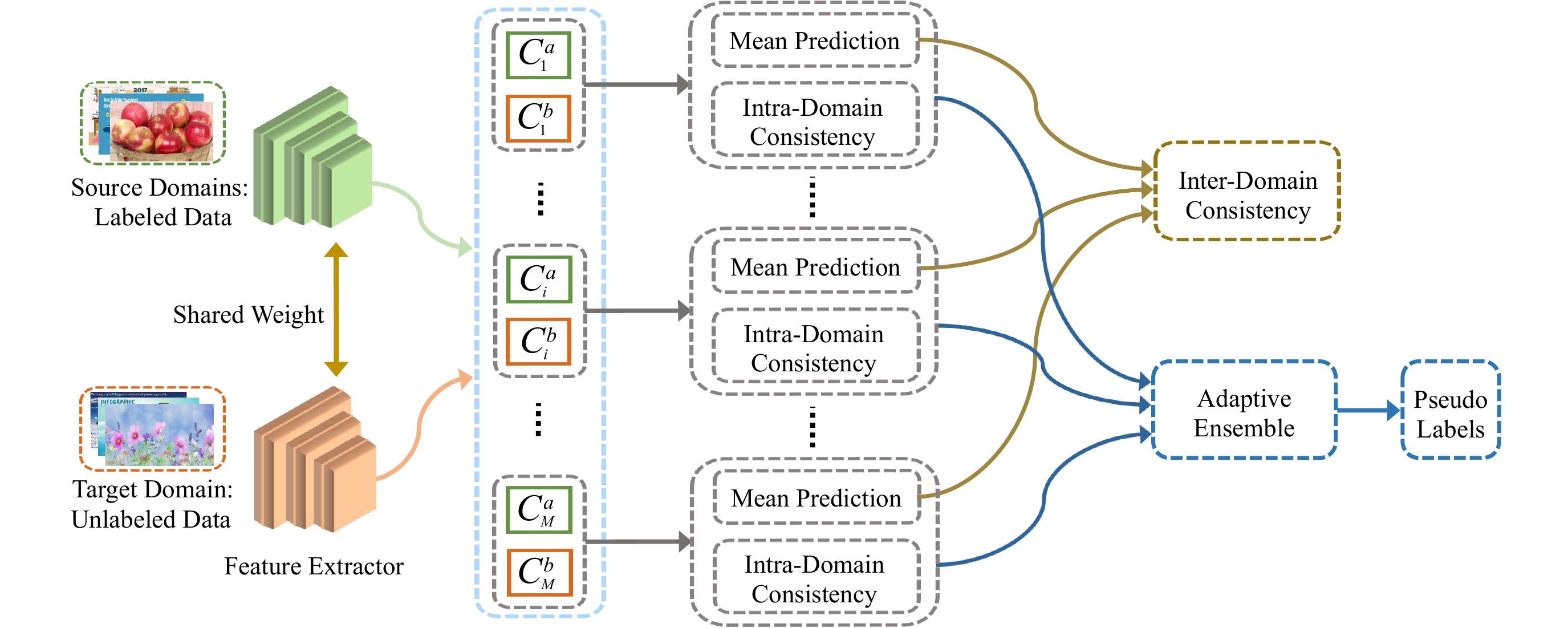}\\
  \caption{The architecture of our proposed MUDA network CRMA: We adapt from multiple labeled \textit{Source Domains} to one unlabeled \textit{Target Domain}. For each source domain, we train two domain-specific classifiers and employ \textit{Intra-Domain Consistency} for intra-domain alignment between the source and target domains. We design \textit{Inter-Domain Consistency} to fuse multiple \textit{Mean Predictions} for jointly aligning across all domains. We also design an \textit{Adaptive Ensemble} strategy based on \textit{Intra-Domain Consistency} which predicts \textit{Pseudo Labels} adaptively for handling negative transfer. $C_i^a$ and $C_i^b$ denote the classifier pair for the i-th source domain, while $M$ is the number of source domains. Best viewed in color. 
  }
  \label{fig:model}
\end{figure}

To address the aforementioned issues, we propose an end-to-end trainable network that exploits Consistency Regularization for unsupervised Multi-source domain Adaptive classification (CRMA). CRMA performs both intra-domain alignment and inter-domain alignment as illustrated in Fig. \ref{fig:model}. On top of training a feature extractor and a pair of classifiers for each source domain \cite{Saito2017Maximum}, we compute the intra-domain consistency of target predictions for each classifier pair and adopt a min-max optimization strategy for domain-specific feature alignment between the target domain and each source domain. In addition, we maximize the inter-domain consistency as computed from the target predictions of different domain-specific classifiers to boost the feature space alignment across all domains. To facilitate model selection and avoid the \textit{negative transfer} \cite{pan2009survey} issue, we design an adaptive self-training strategy that treats the intra-domain consistency as a confidence indicator and uses it to fuse the domain-specific predictions to generate pseudo labels and refine the network.
Despite employing multiple classifiers, the proposed CRMA introduces little overheads in model size and computational complexity as the feature extractor is shared which carries most weights in a typical CNN architecture. Extensive experiments show that CRMA outperforms state-of-the-art methods consistently with clear margins across multiple MUDA datasets.

The contributions of this work can be summarized in three aspects. First, we propose an end-to-end trainable MUDA network that leverages intra-domain alignment and inter-domain alignment for effective adaptation from multiple source domains to one target domain. Second, we develop an adaptive self-training strategy that serves the function of model selection and tackles \textit{negative transfer} effectively. Third, extensive experiments show that our method achieves superior domain adaptation performance consistently across multiple MUDA datasets.

\section{Related Works}

\textbf{Single-source Unsupervised Domain Adaptation (SUDA).} SUDA aims to learn a model well-performing on the target domain given a labeled source domain and an unlabeled target domain. It is usually achieved by discrepancy-based methods \cite{tzeng2014deep, han2020visual, yao2020discriminative}, adversarial learning \cite{ganin2015, zuo110challenging, rahman2020correlation, huang2020contextual, zhang2021detr}, and self-training methods \cite{liang2019exploring}. Tzeng et al. \cite{tzeng2014deep} first proposed Maximum Mean Discrepancy (MMD) to measure the distance between domain distributions. Han et al. \cite{han2020visual} designed a modified $\mathcal{A}$-distance to preserve the internal structures for target domain examples during domain adaptation. Yao et al. \cite{yao2020discriminative} introduced a unified framework that incorporates discriminative embedding constraints and distribution alignment. Ganin et al. \cite{ganin2015} first utilized a domain discriminator to align feature representations with adversarial learning. Zuo et al. \cite{zuo110challenging} applied different strategies to easy and tough examples to improve the domain adaptation performance. Rahman et al. \cite{rahman2020correlation} combined correlation alignment with adversarial learning to tackle the domain adaptation and domain generalization problems. Liang et al. \cite{liang2019exploring} proposed to leverage the uncertainty of pseudo labels to achieve optimal feature transformation. Several studies address category-level feature alignment using dual task-specific classifiers \cite{Saito2017Maximum} and prototypical networks \cite{pan2019transferrable}. SUDA methods usually suffer from sub-optimal performance when directly applied to MUDA tasks since different source domains might have different levels of similarity to the target domain. \textcolor{black}{Our proposed method extends \cite{Saito2017Maximum} but works under a more complicated multi-source setting. The key difference is that the feature alignment needs to be performed between the target domain and multiple source domains that have different similarities with the target domain.}


\textbf{Multi-source Unsupervised Domain Adaptation (MUDA).} MUDA aims to adapt from multiple labeled source domains to one unlabeled target domain. Yang et al. \cite{Yang2007Cross} first introduced the output ensemble of source-domain classifiers for tuning the target-domain categorization model. This idea was later extended by several shallow models via feature representation \cite{Qian2011A} and a combination of pre-learned classifiers \cite{Xu2012Multi} under certain theoretical supports \cite{Liu2016Structure,hoffman2018}. In recent years, deep learning-based approaches have been developed to tackle the MUDA challenge by extracting domain invariant representations among all domains. Xu et al. \cite{Xu2018Deep} presented a deep cocktail network (DCTN) that adopts adversarial learning and employs perplexity scores for target prediction voting. Zhao et al. \cite{zhao2018} introduced a multi-source domain adversarial network (MDAN) that combines the gradient of domain classifiers for bound optimization. M\textsuperscript{3}SDA \cite{peng2019moment} applies moment matching to align the feature representations of multiple domains dynamically. Zhao et al. \cite{zhao2020} proposed a multi-source distilling domain adaptation (MDDA) network to handle different similarities between multiple source domains and the target domain by generating a weighted ensemble of multiple target predictions. Wang et al. \cite{wang2020learning} applied prototypical networks and knowledge graph for knowledge aggregation.

Existing adversarial learning methods focus on learning domain-invariant representation across all domains, while the task-specific decision boundaries among different classes are neglected. Moreover, when aggregating multiple target predictions, existing methods rely on pre-training \cite{Xu2018Deep, peng2019moment} or require multi-stage training \cite{zhao2020}, which are complicated and sub-optimal. In this work, we design an end-to-end framework that leverages the consistency of target predictions for domain-specific and cross-domain feature alignment. We also incorporate model selection into the training process through a novel self-training mechanism.

\begin{figure}[t]
  \centering
  \includegraphics[width=\textwidth]{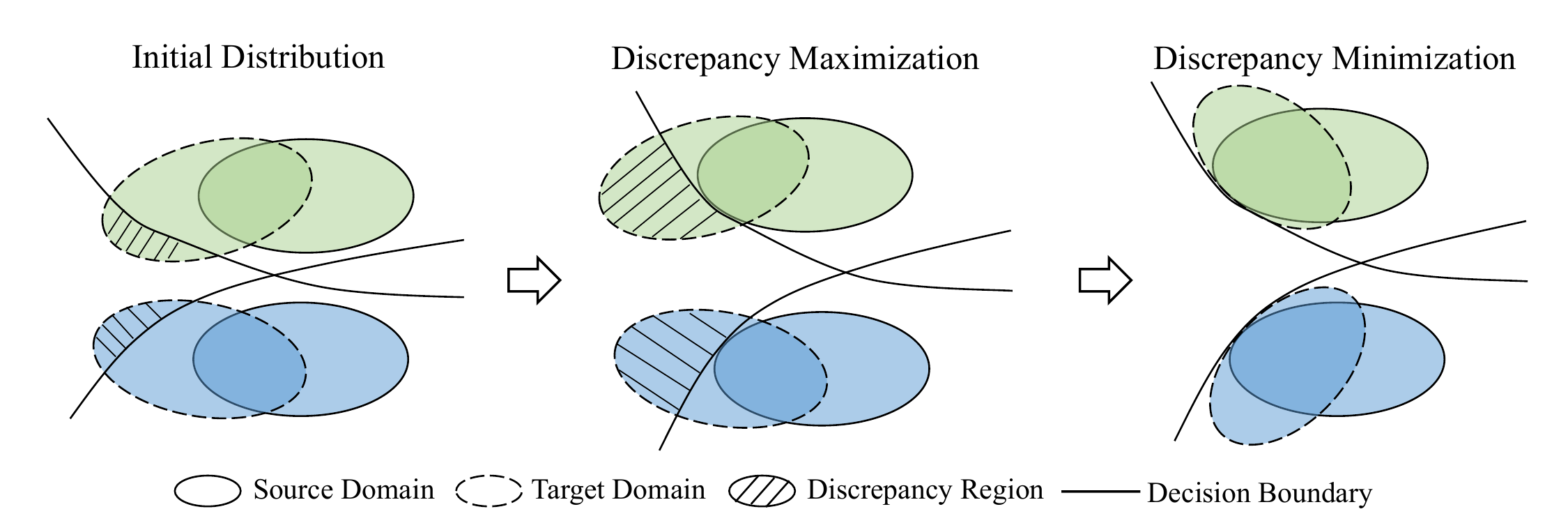}\\
  \caption{\textcolor{black}{
  Illustration of feature alignment with maximum classifier discrepancy (MCD) \cite{Saito2017Maximum}: The discrepancy between two classifiers (for each source domain) is first maximized to detect target samples that are misaligned with the source domain. It is then minimized to guide the feature extractor to learn domain-invariant feature representations for the source and target domains. Green and blue colors denote two different classes. Best viewed in color.}
  }
  \label{fig:mcd}
\end{figure}

\section{Methods}
\subsection{System Overview}
Suppose we have \textit{M} labeled source domains \textit{{$\{\{X_{1}, Y_1\}, \{X_{2}, Y_2\}, ..., \{X_{M}, \\
Y_M\}\}$}}
and one unlabeled target domain \textit{\{$X_T$\}}. We train a model that consists of a feature extractor \textit{$F$} shared across all domains as well as two domain-specific classifiers for each source domain which are denoted by \textit{$\{(C_1^a, C_1^b),(C_2^a, C_2^b),..., \\
(C_M^a, C_M^b)\}$} as shown in Fig. \ref{fig:model}. Given an input image \textit{$x$}, we use \textit{$p(y|x)$} to denote the prediction of the model, e.g. we have \textit{$p_m^a(y|x) = C_m^a(F(x))$} for the first classifier of the \textit{m}-th source domain. The goal of the MUDA task is to maximize the performance on the target domain.

\textcolor{black}{We design Intra-Domain Alignment (IntraDA), Inter-Domain Alignment (InterDA), and Adaptive Self-Training (AST) for MUDA. Inspired by the SUDA method MCD \cite{Saito2017Maximum}, IntraDA exploits a similar min-max optimization strategy that employs two classifiers to align source and target features. As illustrated in Fig. \ref{fig:mcd}, the two classifiers are first trained to maximize their discrepancy to detect target samples that are misaligned with the source domain. The discrepancy is then minimized to guide the feature extractor to learn domain-invariant representations that can better classify those misaligned target samples. Different from adversarial methods that perform domain-level alignment, it performs class-level alignment with class-specific decision boundaries of the two classifiers, which can better handle target samples around the decision boundaries.}


\begin{figure}[t]
  \centering
  \includegraphics[width=\textwidth]{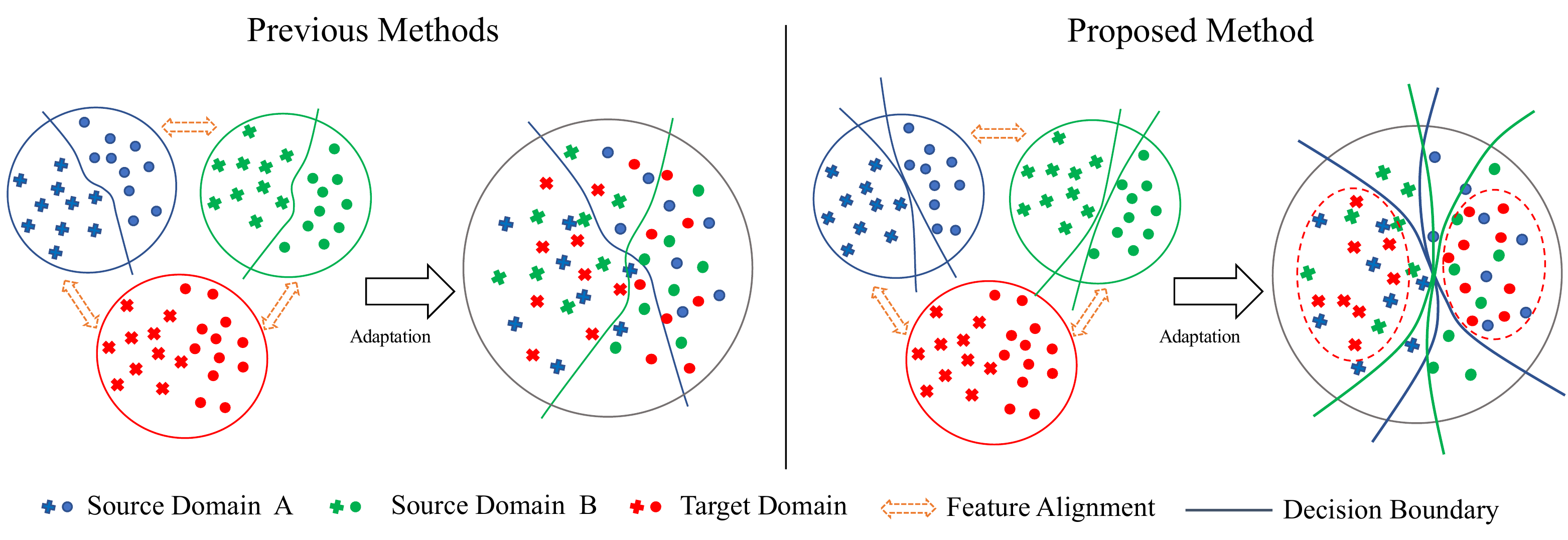}\\
  \caption{\textcolor{black}{
  Illustration of feature alignment with existing MUDA methods and the proposed CRMA: Existing MUDA methods align features at domain level without considering class-specific decision boundaries and tend to misclassify target samples lying around the decision boundaries of source domains. With IntraDA and InterDA, the proposed CRMA aligns target features to the overlapped regions of different source domains, which enables more accurate target sample classification. Best viewed in color.}
  }
  \label{fig:compare}
\end{figure}

\textcolor{black}{IntraDA aligns the target domain with each source domain individually but cannot handle MUDA well when multiple source domains of different distributions are present. We design InterDA to capture the synergy of multiple source domains while jointly aligning with the target domain. Specifically, InterDA computes inter-domain consistency that measures how the predictions of multiple domain-specific classifiers agree with each other. By maximizing this consistency with a similar mechanism adopted in IntraDA, InterDA encourages consistent target predictions from domain-specific classifiers. As illustrated in Fig. \ref{fig:compare}, IntraDA and InterDA move target features to the overlapped regions of different source domains, which can better classify target samples lying around the decision boundaries. As a comparison, existing adversarial MUDA methods align features without considering class-specific decision boundaries, which tend to misclassify those target samples.}

\textcolor{black}{Another feature of MUDA is that different source domains have different similarities to the target domain and should have different weights while performing the alignment. While some source domains have very different distributions from the target domain, feature alignment could even suffer from \textit{negative transfer} that affects target performance negatively. The proposed Adaptive Self-Training (AST) generates pseudo labels (for target samples) by assigning authorities to domain-specific classifiers adaptively. The idea is to assign higher authorities to more confident classifiers which mitigates the negative effects of low-confidence classifiers effectively. In CRMA, we determine the classifier authorities by using the intra-domain consistency which provides a good measure of prediction confidence for different domain-specific classifiers. More details of the three designs are to be discussed in the ensuing subsections.}


\subsection{Intra-Domain Alignment}
The shared feature extractor and domain-specific classifiers are first trained with source-domain samples to obtain discriminative features by using softmax cross entropy loss:

\begin{equation} \label{eq: 1}
\begin{aligned}
     \underset{F, C}{\text{min}} \;  L_{src} = - \sum_{m=1}^M \mathbb{E}_{(x_m, y_m)\sim(X_m, Y_m)}\sum_{k=1}^K \mathbbm{1}_{[k=y_m]}\log{p(y|x_m)}
\end{aligned}
\end{equation}

\noindent where $K$ denotes the number of classes. Subsequently, we compute the intra-domain consistency by summing up the discrepancy between the target predictions of the two domain-specific classifiers and perform intra-domain consistency minimization by fixing the feature extractor \textit{F} and training the domain-specific classifiers \textit{C}. It adjusts the decision boundaries of the classifiers so that they could detect target samples misaligned with the source domains (i.e. where the classifier pairs disagree with each other). We follow the practice in \cite{Saito2017Maximum} to perform this update with the source data classification loss. The objective can be described by:

\begin{equation} \label{eq: 2}
    \underset{C}{\text{min}} \; L_{src} - L_{intra}
\end{equation}
where
\begin{equation} \label{eq: 3}
     L_{intra} = \mathbb{E}_{x_t\sim X_t} \sum_{m=1}^M d(p_m^a(y|x_t), p_m^b(y|x_t))
\end{equation}
$L_{intra}$ stands for the intra-domain consistency loss and \textit{d} denotes the discrepancy in the form of L1 loss that measures the distance between two prediction probability vectors:

\begin{equation} \label{eq: 4}
    d(p, q) = \frac{1}{K} \sum_{k=1}^{K} |p_{k} - q_{k}|
\end{equation}

Next, we conduct the consistency maximization step which works in an adversarial manner with the consistency minimization step by fixing the classifiers and updating the shared feature extractor. This aims to train the feature extractor to generate domain-invariant features for the target domain and each source domain. In practice, we combine this update with the inter-domain alignment to be discussed in the next subsection to simplify the training process. 

\subsection{Inter-Domain Alignment}
While IntraDA aligns the target domain and individual source domains, InterDA aims to learn domain-invariant representations across all domains. It works in a similar fashion as IntraDA by maximizing the inter-domain consistency of the target predictions. The inter-domain consistency $L_{inter}$ is calculated by summing up the discrepancy among the mean predictions of all domain-specific classifier pairs: 

\begin{equation} \label{eq: 5}
     L_{inter} = \mathbb{E}_{x_t\sim X_t} \sum_{m=1}^{M}\sum_{n=m+1}^{M} d(\hat{p}_m, \hat{p}_n)
\end{equation}
where
\begin{equation}  \label{eq: 6}
    \hat{p}_m = (p_m^a(y|x_t) + p_m^b(y|x_t))/2
\end{equation}
where $\hat{p}_m$ stands for the mean prediction of the classifier pair corresponds to the m-th domain. We then combine local and inter-domain consistency losses and minimize the loss by fixing the classifiers and updating the feature extractor:
\begin{equation} \label{eq: 7}
    \underset{F}{\text{min}} \;  L_{intra} + \alpha L_{inter}
\end{equation}
where $\alpha$ is the loss ratio for the inter-domain consistency loss, which is a hyper-parameter of the proposed network.

InterDA works under a similar principle as IntraDA that it guides the feature extractor to generate target domain features that align with all source domains by boosting the inter-domain consistency of the target predictions from different source domains. Note that we do not perform the inter-domain consistency minimization as done in IntraDA because the classifier pairs correspond to different source domains naturally form different decision boundaries as labeled data from different source domains are used for their training, which leads to inconsistency in the target predictions.

\subsection{Adaptive Self-Training}
On top of consistency-based feature alignment, we design a self-training mechanism to generate pseudo labels for target samples by weighting the target predictions of different domain-specific classifier pairs adaptively. We use the intra-domain consistency as an indicator of prediction confidence since better alignments between the source and target domains lead to a smaller discrepancy and thus, higher confidence in the target predictions. Specifically, the pseudo label $P(y|x_t)$ is generated as follows:
\begin{equation} \label{eq: 8}
    P(y|x_t) =  \sum_{m=1}^{M} \frac{w_m}{\sum_{n=1}^{M}w_n} \hat{p}_m
\end{equation}
where
\begin{equation} \label{eq: 9}
    w_m = 1 / (L_{intra}^m + \lambda \overline{L}_{intra}^m)
\end{equation}
\textcolor{black}{where $w_m$ is the weight of the m-th source domain which is computed based on the intra-domain consistency loss $L_{intra}^m$ and the mean of $L_{intra}^m$ over all seen training samples denoted by $\overline{L}_{intra}^m$. As $L_{intra}^m$ captures the prediction confidence on the current training sample, $\overline{L}_{intra}^m$ captures the mean prediction confidence which is affected by the similarity between the m-th source domain and the target domain. Here $\overline{L}_{intra}^m$ acts as a regularization factor which mitigates large fluctuations of $w_m$ and the modulation strength is controlled by $\lambda$.}

After obtaining the pseudo label, we update the model using KL divergence loss \cite{kullback1951information}:
\begin{equation} \label{eq: 10}
\begin{aligned}
    \underset{F, C}{\text{min}} \; L_{AST} = \mathbb{E}_{x_t\sim X_t} \sum_{m=1}^M  \beta & (D_{KL}(p_m^a(y|x_t) || P(y|x_t))\\
    + & D_{KL}(p_m^b(y|x_t) || P(y|x_t)))
\end{aligned}
\end{equation}
where
\begin{equation} \label{eq: 11}
    \beta = \min{(\overline{L}_{intra}^m)} \sum_{m=1}^M w_m 
\end{equation}

\textcolor{black}{Parameter $\beta$ is the weight of the adaptive self-training loss $L_{AST}$, which aims to suppress the impact of the less confident pseudo labels in the self-training process. It is computed by summing up $w_m$ over $M$ source domains and multiplying with the minimum of $\overline{L}_{intra}^m$ among source domains. As $w_m$ is a confidence indicator of individual target prediction, the sum of $w_m$ captures the overall prediction confidence from all source domains on the current target sample. While $\overline{L}_{intra}^m$ captures the mean confidence over all samples, $\min{(\overline{L}_{intra}^m)}$ is the mean prediction confidence by the most confident source domain which is a modulator for $\beta$. Under such designs, the generated pseudo labels with less overall confidence will be assigned with smaller weights while computing $L_{AST}$ and thus have smaller impacts on the learning process.}

\renewcommand{\algorithmicrequire}{\textbf{Input:}}
\renewcommand{\algorithmicensure}{\textbf{Output:}}

\begin{algorithm}[t]
\caption{Consistency-Regularized Self-Training for multi-source domain adaptation (CRMA)} 
\label{alg: alg}
\begin{algorithmic}[1]
\REQUIRE labeled source domains \textit{{$\{\{X_{1}, Y_1\}, \{X_{2}, Y_2\}, ..., $}} \textit{{$\{X_{M}, Y_M\}\}$}}, an unlabeled target domain \textit{$X_T$}. Feature extractor $F$ and classifiers $C$.
\ENSURE Trained feature extractor \textit{$F'$} and classifiers \textit{$C'$}.
\FOR{$iteration=1,2,\ldots$}
\STATE Sample $\{x_m, y_m\}_{m=1}^M$ from the source domains and $x_t$ from the target domain
\STATE Update feature extractor $F$ and classifiers $C$ using Eq. \ref{eq: 1} with source data
\STATE Perform intra-domain consistency minimization by updating $C$ with Eq. \ref{eq: 2}
\STATE Compute intra-domain consistency $L_{intra}$ with Eq. \ref{eq: 3}
\STATE Compute intra-domain consistency $L_{inter}$ with Eq. \ref{eq: 5}
\STATE Perform consistency maximization by updating $F$ with Eq. \ref{eq: 7}
\STATE Compute pseudo labels $P(y|x)$ with Eq. \ref{eq: 8}
\STATE Update $F$ and $C$ using $P(y|x)$ with Eq. \ref{eq: 10}
\ENDFOR
\RETURN $F'=F$, $C'=C$
\end{algorithmic} 
\end{algorithm}

\subsection{Network Training}
Algorithm \ref{alg: alg} summarizes the CRMA training process. In each training iteration, we randomly pick training samples from each source domain and the target domain and train a pair of domain-specific classifiers by using the sampled source samples (with labels). The intra-domain consistency is computed according to the target predictions of domain-specific classifier pairs, while the mean prediction is derived from the classifier pairs' predictions to compute the inter-domain consistency. intra-domain consistency minimization is performed to allow classifiers to detect misaligned target samples and both local and inter-domain consistency are maximized to achieve domain-specific and domain-agnostic feature alignment. For AST, pseudo labels of target-domain samples are predicted by taking the weighted average of the mean predictions based on intra-domain consistency and used to update the whole network. In evaluation, we average the probability vectors of all classifiers to generate the final prediction.

\section{Experiments}

\subsection{Datasets}
We compare our CRMA with state-of-the-art methods over three public datasets as listed:

\textbf{Digits-5} contains images of digits from five different visual domains including handwritten images in MNIST \cite{lecun1998gradient} and USPS \cite{hull1994database}, combined images in MNIST-M \cite{ganin2014unsupervised}, street images in SVHN \cite{netzer2011reading}, and synthetic images in SYN \cite{ganin2014unsupervised}. For fair comparisons, we follow the same training-testing split as in \cite{peng2019moment}. 

\textbf{DomainNet} \cite{peng2019moment} is a recent large-scale domain adaptation dataset. Images in DomainNet are collected from the Internet and categorized into six domains including clipart, infograph, painting, quickdraw, real, and sketch. The total number of images in this dataset is about 600,000 and each domain has 345 categories. 

\textbf{PACS} \cite{li2017deeper} contains images from four domains which are art painting, cartoon, photo, and sketch. Each domain contains objects from 7 categories.

\subsection{Experimental Setups}

We compare CRMA with state-of-the-art MUDA methods including MDAN \cite{zhao2018a}, DCTN \cite{Xu2018Deep}, M\textsuperscript{3}SDA \cite{peng2019moment}, MDDA \cite{zhao2020}, and LtC-MSDA \cite{wang2020learning}. Being an emerging research area, MUDA has relatively small literature so we also compare with SUDA methods DAN \cite{long2015learning}, DANN \cite{ganin2015}, ADDA \cite{Tzeng2017Adversarial}, and MCD \cite{Saito2017Maximum} for more comprehensive evaluations. For the compared SUDA methods, both single-best and source-combined setups are evaluated, where the former adapts each source domain separately and reports the best model accuracy on the target domain while the latter combines all source domains as a single source domain for adaptation and evaluation. 

\begin{table}[h]
  \caption{The training setups for the three studied datasets: The \textit{Batch size} denotes the number of training samples in each mini-batch drawn from each domain during training.
  }
  \label{experiment setups}
  \centering
  \scalebox{0.8}{
  \begin{tabular}{c|c|c|c|c|c}
    \toprule
    Dataset & Backbone & Image size & \makecell{Batch \\ size} & \makecell{Learning \\ rate} & Epoch \\
    \midrule
    Digits-5 & Lenet \cite{lecun1998gradient} & $32\times 32$ & 128 & $1\times 10^{-3}$ & 50 \\
    DomainNet & Resnet-101 \cite{he2016deep} & $224\times 224$ & 16 & $1\times 10^{-3}$ & 10 \\
    PACS & Resnet-18 \cite{he2016deep} & $224\times 224$ & 16 & $1\times 10^{-3}$ & 100 \\
    \bottomrule
  \end{tabular}
  }
\end{table}

In the experiments, we use the same backbone networks as the comparing methods on each of the three datasets. Table \ref{experiment setups} shows the network architectures and the training parameters for the experiments. Specifically, Lenet \cite{lecun1998gradient} consists of three convolutional layers and three fully connected layers, and the input channels for the three fully connected layers are 8192, 3072, and 2048, respectively. For Resnet-101 and Resnet-18 \cite{he2016deep}, we use two fully connected layers following the convolution blocks with input channels of (2048, 1000) and (512, 512), respectively. In all our experiments, we treat the convolutional layers as the feature extractor $F$ and fully connected layers as the classifier $C$. The Lenet model is trained from scratch. For ResNet-18 and ResNet-101, we follow \cite{wang2020learning} and load the checkpoints pre-trained on ImageNet \cite{deng2009imagenet} as the feature extractor. Besides, we adopt the practice in \cite{zhu2019aligning} and assign a smaller learning rate to the pre-trained feature extractors during training, which is 1/10 of the base learning rate. To speed up the training process for the dataset DomainNet, we apply the cosine annealing scheduling \cite{loshchilov2016sgdr} to adjust the learning rate. $\alpha$ is set to 0.5 and $\lambda$ is set to 0.1 for all the experiments. For each experiment, we conduct five random runs and report the average performance. 

\begin{table*}[t]
  \centering
  \caption{Comparing CRMA with the state-of-the-art on Digits-5 (in classification accuracy \%).}
  \label{digits5 table}
  \scalebox{0.75}{
  \begin{tabular}{c|c|c|c|c|c|c|c}
    \toprule
    Standards & Methods & MNIST-M & MNIST & USPS & SVHN & SYN & Avg \\
    \midrule
    \multirow{4}{4em}{Single Best}
    & Source Only & 59.2$\pm$0.6 & 97.2$\pm$0.6 & 84.7$\pm$0.8 & 77.7$\pm$0.8 & 85.2$\pm$0.6 & 80.8 \\
    & DAN \cite{long2015learning}  & 63.9$\pm$0.7 & 96.3$\pm$0.5 & 94.2$\pm$0.9 & 62.5$\pm$0.7 & 85.4$\pm$0.8 & 80.4 \\
    & DANN \cite{ganin2015} & 71.3$\pm$0.6 & 97.6$\pm$0.8 & 92.3$\pm$0.9 & 63.5$\pm$0.8 & 85.4$\pm$0.8 & 82.0 \\
    & ADDA \cite{Tzeng2017Adversarial} & 71.6$\pm$0.5 & 97.9$\pm$0.8 & 92.8$\pm$0.7 & 75.5$\pm$0.5 & 86.5$\pm$0.6 & 84.8 \\
    \hline
    \multirow{4}{4em}{Source Combine}
    & Source Only & 63.4$\pm$0.7 & 90.5$\pm$0.8 & 88.7$\pm$0.9 & 63.5$\pm$0.9 & 82.4$\pm$0.6 & 77.7 \\
    & DANN \cite{ganin2015} & 70.8$\pm$0.8 & 97.9$\pm$0.7 & 93.5$\pm$0.8 & 68.5$\pm$0.5 & 87.4$\pm$0.9 & 83.6 \\
    & ADDA \cite{Tzeng2017Adversarial} & 72.3$\pm$0.7 & 97.9$\pm$0.6 & 93.1$\pm$0.8 & 75.0$\pm$0.8 & 86.7$\pm$0.6 & 85.0 \\
    & MCD \cite{Saito2017Maximum}  & 72.5$\pm$0.7 & 96.2$\pm$0.8 & 95.3$\pm$0.7 & 78.9$\pm$0.8 & 87.5$\pm$0.7 & 86.1 \\
    \hline
    \multirow{6}{4em}{Multi-Source} 
    & MDAN \cite{zhao2018a}  & 69.5$\pm$0.3 & 98.0$\pm$0.9 & 92.4$\pm$0.7 & 69.2$\pm$0.6 & 87.4$\pm$0.5 & 83.3 \\
    & DCTN \cite{Xu2018Deep}  & 70.5$\pm$1.2 & 96.2$\pm$0.8 & 92.8$\pm$0.3 & 77.6$\pm$0.4 & 86.8$\pm$0.8 & 84.8 \\
    & M\textsuperscript{3}SDA \cite{peng2019moment} & 72.8$\pm$1.1 & 98.4$\pm$0.7 & 96.1$\pm$0.8 & 81.3$\pm$0.9 & 89.6$\pm$0.6 & 87.7 \\
    & MDDA \cite{zhao2020}  & 78.6$\pm$0.6 & 98.8$\pm$0.4 & 93.9$\pm$0.5 & 79.3$\pm$0.8 & 89.7$\pm$0.7 & 88.1 \\
    & LtC-MSDA \cite{wang2020learning} & 85.6$\pm$0.8 & \textbf{99.0}$\pm$0.4 & \textbf{98.3}$\pm$0.4 & 83.2$\pm$0.6 & 93.0$\pm$0.5 & 91.8 \\
    & \textbf{CRMA} & \textbf{94.5}$\pm$0.4 & \textbf{99.0}$\pm$0.1 & 98.0$\pm$0.3 & \textbf{85.6}$\pm$1.0 & \textbf{94.6}$\pm$0.1 & \textbf{94.3} \\
    \bottomrule
  \end{tabular}
  }
\end{table*}

\subsection{Experimental Results}
For each dataset evaluated, we take turns to put each domain as the target domain and the rest as the source domains.

\begin{table*}[t]
 \centering
  \caption{Comparing CRMA with the state-of-the-art on DomainNet (in classification accuracy \%).}
  \label{domainnet table}
  \scalebox{0.65}{
  \begin{tabular}{c|c|c|c|c|c|c|c|c}
    \toprule
    Standards & Methods & Clipart & Infograph & Painting & Quickdraw & Real & Sketch & Avg \\
    \midrule
    \multirow{4}{4em}{Single Best} 
    & Source Only & 39.6$\pm$0.6 & 8.2$\pm$0.8 & 33.9$\pm$0.6 & 11.8$\pm$0.7 & 41.6$\pm$0.8 & 23.1$\pm$0.7 & 26.4 \\
    & DANN \cite{ganin2015} & 37.9$\pm$0.7 & 11.4$\pm$0.9 & 33.9$\pm$0.6 & 13.7$\pm$0.6 & 41.5$\pm$0.7 & 28.6$\pm$0.6 & 27.8 \\
    & ADDA \cite{Tzeng2017Adversarial} & 39.5$\pm$0.8 & 14.5$\pm$0.7 & 29.1$\pm$0.8 & 14.9$\pm$0.5 & 41.9$\pm$0.8 & 30.7$\pm$0.7 & 28.4\\
    & MCD \cite{Saito2017Maximum}  & 42.6$\pm$0.3 & 19.6$\pm$0.8 & 42.6$\pm$1.0 & 3.8$\pm$0.6 & 50.5$\pm$0.4 & 33.8$\pm$0.9 & 32.2 \\
    \hline
    \multirow{4}{4em}{Source Combine} 
    & Source Only & 47.6$\pm$0.5 & 13.0$\pm$0.4 & 38.1$\pm$0.5 & 13.3$\pm$0.4 & 51.9$\pm$0.9 & 33.7$\pm$0.5 & 32.9 \\
    & DANN \cite{ganin2015} & 45.5$\pm$0.6 & 13.1$\pm$0.7 & 37.0$\pm$0.7 & 13.2$\pm$0.8 & 48.9$\pm$0.7 & 31.8$\pm$0.6 & 32.6 \\
    & ADDA \cite{Tzeng2017Adversarial} & 47.5$\pm$0.8 & 11.4$\pm$0.7 & 36.7$\pm$0.5 & 14.7$\pm$0.5 & 49.1$\pm$0.8 & 33.5$\pm$0.5 & 32.2 \\
    & MCD \cite{Saito2017Maximum}  & 54.3$\pm$0.6 & 22.1$\pm$0.7 & 45.7$\pm$0.6 & 7.6$\pm$0.5 & 58.4$\pm$0.7 & 43.5$\pm$0.6 & 38.5 \\
    \hline
    \multirow{6}{4em}{Multi-Source} 
    & MDAN \cite{zhao2018a} & 52.4$\pm$0.6 & 21.3$\pm$0.8 & 46.9$\pm$0.4 & 8.6$\pm$0.6 & 54.9$\pm$0.6 & 46.5$\pm$0.7 & 38.4 \\
    & DCTN \cite{Xu2018Deep} & 48.6$\pm$0.7 & 23.5$\pm$0.6 & 48.8$\pm$0.6 & 7.2$\pm$0.5 & 53.5$\pm$0.6 & 47.3$\pm$0.5 & 38.2 \\
    & M\textsuperscript{3}SDA \cite{peng2019moment} & 58.6$\pm$0.5 & 26.0$\pm$0.9 & 52.3$\pm$0.6 & 6.3$\pm$0.6 & 62.7$\pm$0.5 & 49.5$\pm$0.8 & 42.6 \\
    & MDDA \cite{zhao2020} & 59.4$\pm$0.6 & 23.8$\pm$0.8 & 53.2$\pm$0.6 & 12.5$\pm$0.6 & 61.8$\pm$0.5 & 48.6$\pm$0.8 & 43.2 \\
    & LtC-MSDA \cite{wang2020learning} & 63.1$\pm$0.5 & \textbf{28.7}$\pm$0.7 & \textbf{56.1}$\pm$0.5 & 16.3$\pm$0.5 & 66.1$\pm$0.6 & 53.8$\pm$0.6 & 47.4 \\
    & \textbf{CRMA} & \textbf{67.6}$\pm$0.6 & 25.3$\pm$0.4 & 55.1$\pm$0.3 & \textbf{18.4}$\pm$0.5 & \textbf{66.9}$\pm$0.3 & \textbf{56.0}$\pm$0.3 & \textbf{48.2} \\
    \bottomrule
  \end{tabular}
  }
\end{table*}

\begin{table}[t]
  \centering
  \caption{Comparing CRMA with the state-of-the-art on PACS (in classification accuracy \%). (A: art painting, C: cartoon, S: sketch, P: photo)}
  \label{pacs table}
  \scalebox{0.88}{
  \begin{tabular}{c|c|c|c|c|c|c}
    \toprule
    Standards & Methods & A & C & S & P & Avg \\
    \midrule
    \multirow{3}{4em}{Single Best}
    & Source Only & 68.2 & 60.0 & 61.8 & 95.2 & 71.3 \\
    & ADDA \cite{Tzeng2017Adversarial} & 75.9 & 80.7 & 66.4 &93.0 & 79.0 \\
    & MCD \cite{Saito2017Maximum} & 81.2 & 84.5 & 65.9 & 96.9 & 82.1 \\
    \hline
    \multirow{3}{4em}{Source Combine}
    & Source Only & 82.9 & 76 & 66.4 & 93.2	& 79.6 \\
    & ADDA \cite{Tzeng2017Adversarial} & 87 & 83.9 & 73.7 & 94.7 & 84.8\\
    & MCD \cite{Saito2017Maximum} & 88.2 & 85.2 & 61.0 & 97.2 & 82.9 \\
    \hline
    \multirow{6}{4em}{Source Combine}
    & MDAN \cite{zhao2018a} & 83.5 & 86.7 & 71.8 & 94.5 & 84.7 \\
    & DCTN \cite{Xu2018Deep} & 84.7 & 86.7 & 71.8 & 95.6 & 84.7 \\
    & M\textsuperscript{3}SDA \cite{peng2019moment} & 84.2 & 85.7 & 74.6 & 94.5 & 84.7\\
    & MDDA \cite{zhao2020} & 86.7 & 86.2 & 77.6 & 93.9 & 86.1 \\
    & LtC-MSDA \cite{wang2020learning} & 90.2 & 90.5 & \textbf{81.5} & 97.2 & 89.9 \\
    & \textbf{CRMA} & \textbf{91.5} & \textbf{92.3} & 80.9 & \textbf{97.7} & \textbf{90.6} \\
    \bottomrule
  \end{tabular}
  }
\end{table}

Table \ref{digits5 table} shows the experimental results on the Digits-5 dataset. We can see that CRMA achieves an average classification accuracy of 94.3\%, which is 2.5\% higher than the state-of-the-art by LtC-MSDA \cite{wang2020learning}. In addition, CRMA obtains higher or comparable accuracy when all domains are used as the target domains. In particular, when transferring knowledge from other domains to MNIST-M, a significant improvement of 8.9\% is achieved. From the experiments, we observe that the source domains have relatively balanced contributions when MNIST-M is the target domain, which indicates that each source domain shares fair similarity with the target. This implies that our method is especially beneficial in gathering useful information from different source domains, which also explains why MUDA often outperforms SUDA in a multi-source setup. CRMA also achieves an accuracy gain of 2.4\% on SVHN, which is known as the most difficult target domain due to its different similarities to the source domains. The superior accuracy indicates that CRMA is capable of extracting useful knowledge in a noisy environment.

Table \ref{domainnet table} shows the comparison over DomainNet which is a more challenging dataset due to its large number of categories and significant domain shift. CRMA achieves an average accuracy of 48.2\% and performs best in the majority of the target domains. For the challenging target domain \textit{quickdraw} where negative transfer is often observed, CRMA obtains an accuracy of 18.4\% with a clear margin of 2.1\%. 

Table \ref{pacs table} shows the comparison on PACS dataset. It can be seen that CRMA achieves an average accuracy of 90.6\% with a margin of 0.7\% as compared with the state-of-the-art. The performance gain is smaller because this dataset is relatively small and the performance relies heavily on the pre-trained model. In addition, a majority of target domains are very similar to the source domains and the classification accuracy is close to saturation ($>$ 90\%).

\begin{table}[t]
  \caption{Ablation study of CRMA on Digits-5 (in classification accuracy \%). (MM: MNIST-M, MT: MNIST, UP: USPS, SV: SVHN, SY: SYN)
  }
  \label{ablation table}
  \centering
  \scalebox{0.85}{
  \begin{tabular}{ccc|ccccc|c}
    \toprule
    IntraDA & InterDA & AST & MM & MT & UP & SV & SY & Avg \\
    \midrule
    & & & 67.9 & 98.0 & 95.4 & 69.8 & 79.4 & 82.1 \\
    \checkmark & & & 72.9&	98.3&	95.8&	76.9&	84.3&	85.6\\
    & \checkmark & & 75.8&	99.1&	98.2&	75.0&	92.5&	88.1\\
    & & \checkmark & 79.4&	99.1&	98.1&	82.7 &	93.4&	90.5 \\
    \checkmark & \checkmark & & 75.8&	98.3&	97.2&	79.0 &	88.4&	87.7 \\
    \checkmark & & \checkmark & 93.2&	99.0&	97.4&	83.2&	94.1&	93.4 \\
    & \checkmark & \checkmark & 82.5&	99.1&	98.4&	84.1&	94.2&	91.7 \\
    \checkmark & \checkmark & \checkmark & 94.4 &	99.0 &	98.0 &	85.6 &	94.6&	94.3 \\
    \bottomrule
  \end{tabular}
  }
\end{table}

\subsection{Ablation Study}
In this subsection, we conduct ablation studies to analyze the effectiveness and contributions of different components in our proposed CRMA.

\begin{table}[t]
  \caption{Comparing AST with uniform ensemble on dataset Digits-5 (in classification accuracy \%). (MM: MNIST-M, MT: MNIST, UP: USPS, SV: SVHN, SY: SYN)
  }
  \label{ensemble table}
  \centering
  \scalebox{0.9}{
  \begin{tabular}{c|ccccc|c}
    \toprule
    Methods & MM & MT & UP & SV & SY & Avg \\
    \midrule
    Uniform Ensemble & 93.8 & 99.0 & 98.3& 79.0 & 94.3 & 92.9 \\ 
    AST &  94.5 & 99.0 & 98.0 & 85.6 & 94.6 & 94.3\\
    \bottomrule
  \end{tabular}
  }
\end{table}

\begin{figure}[t]
  \centering
  \includegraphics[width=0.6\textwidth]{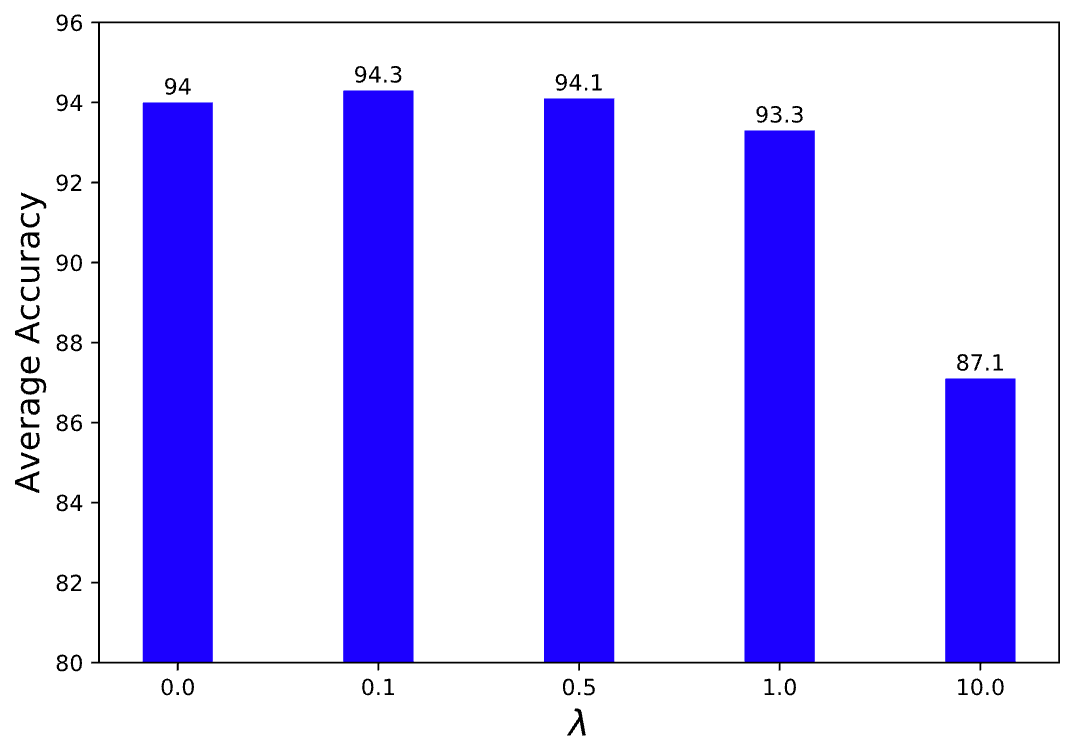}\\
  \caption{\textcolor{black}{Ablation study of of $\lambda$ on Digits-5 dataset. The classification is stable while $\lambda$ lies between 0 and 1 (best performance is obtained while $\lambda=0.1$).} 
  }
  \label{fig:lambda}
\end{figure}

Specifically, CRMA consists of three major components: intra-domain alignment (IntraDA), inter-domain alignment (InterDA), and Adaptive Self-Training (AST). To investigate the contribution of these three components, we conduct a series of ablation studies over Digits-5 by applying relevant losses in network training. We also conduct experiments over the \textit{Source-Only} condition to derive a baseline, where the model is trained by using the labeled source-domain data and applied to the target-domain data without domain adaptation. Table \ref{ablation table} shows experimental results. It is interesting to observe that the \textit{Source-Only} results obtained with our proposed network architecture achieve an average accuracy of 82.1\%, which is higher than both the single-best and source-combine baselines as shown in Table \ref{digits5 table}. This shows that adopting a network architecture of shared feature extractor and domain-specific classifiers leads to more effective network learning and generalization under a multi-source domain adaptation setup. 

We can also observe that when each of the three components is incorporated, there is a clear performance gain as compared to the \textit{Source-Only} baseline. Specifically, incorporating AST achieves the highest accuracy of 90.5\% while incorporating IntraDA introduces the lowest gain, which is expected since IntraDA only addresses domain-specific feature alignment and leads to imbalanced performance across domain-specific classifiers on the target domain. In addition, IntraDA and AST are complementary that incorporating both produces an average accuracy of 93.4\%, largely because the self-training effectively aggregates domain-specific knowledge and closes the performance gap among classifiers. On the other hand, IntraDA and InterDA do not demonstrate clear complementary effects. We observe from the experiments that although InterDA aligns the feature distributions globally, it does not resolve the imbalanced performance of IntraDA well. Furthermore, InterDA is less complementary to AST either as compared to IntraDA as both InterDA and AST address the global alignment problem. Finally as expected, the complete network model with all three components incorporated achieves the best classification accuracy.

\textcolor{black}{We also conduct a sensitivity analysis for the hyper-parameter $\lambda$, a weighting factor in Eq. \ref{eq: 9} that balances individual prediction confidence and the mean confidence. Fig. \ref{fig:lambda} shows the average classification accuracy when $\lambda$ is set to different values (tested on Digits-5). It can be seen that the classification performance is stable while $\lambda$ lies between 0 and 1 and the best accuracy is obtained at $\lambda=0.1$.}

\begin{figure}[t]
  \centering
  \includegraphics[width=\textwidth]{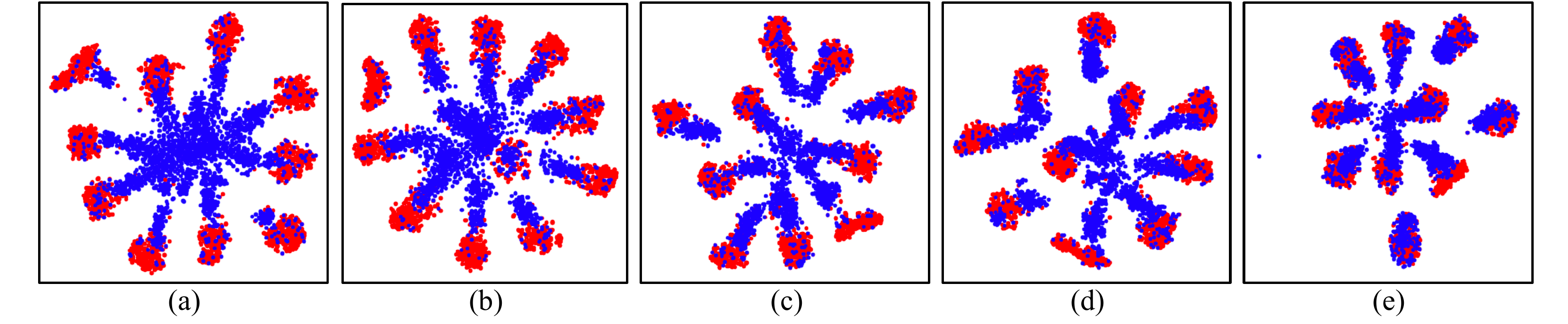}\\
  \caption{t-SNE visualization of feature representations of a source domain (MNIST) and a target domain (MNIST-M) in Digits-5: (a) Source Only with no adaptation, (b) Intra-Domain Alignment, (c) Inter-Domain Alignment, (d) Adaptive Self-Training, (e) IntraDA + InterDA + AST. Red/blue points represent source/target domain. Best viewed in color.
  }
  \label{fig:tsne}
\end{figure}

\begin{figure}[t]
  \centering
  \includegraphics[width= 4 in]{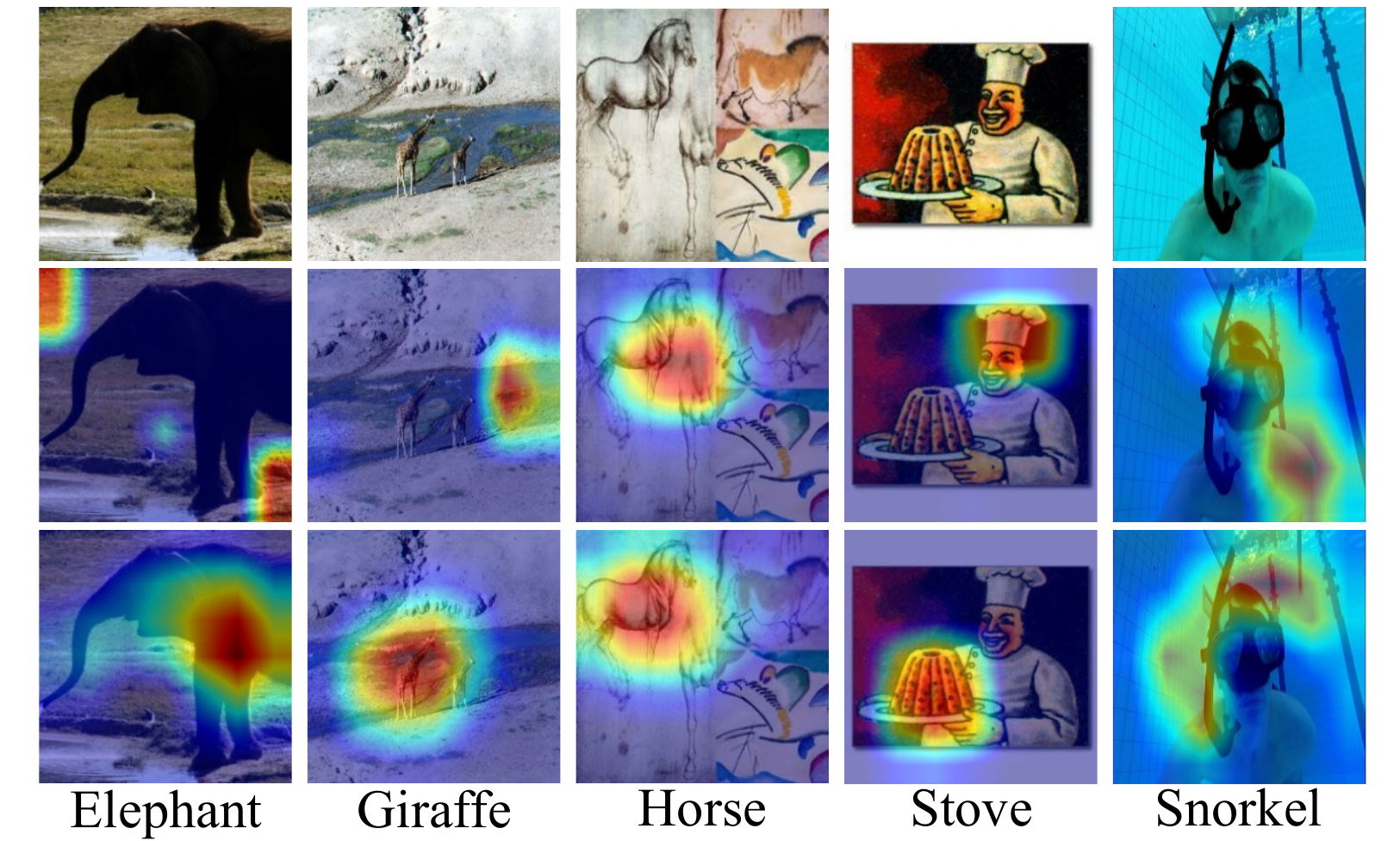}\\
  \caption{Grad-CAM visualization on sample images from PACS and DomainNet datasets: For the sample images in Row 1, Rows 2 and 3 show the corresponding class activation maps with no domain adaptation and with CRMA, respectively. The text below images indicates the class labels. Best viewed in color.}
  \label{fig:cam}
\end{figure}

\subsection{Discussion}
To examine the effectiveness of our adaptive pseudo label generation in AST, we replace our generated pseudo labels with the \textit{Uniform Ensemble} of target predictions (where each classifier has an equal contribution) and benchmark over Digits-5. As Table \ref{ensemble table} shows, AST outperforms the \textit{Uniform Ensemble} by 1.4\% in average classification accuracy. For easier target domains that share good similarity with the source domains, AST achieves similar accuracy as \textit{Uniform Ensemble}. However, for difficult target domain SVHN, AST outperforms \textit{Uniform Ensemble} by a large margin at 6.6\%. Such experimental results further show that our consistency-based ensemble performs model selection effectively by assigning higher authorities (or weights) to the target predictions that are more suitable for adaptation.

We also study how different CRMA components affect the feature distributions via t-SNE visualization \cite{maaten2008visualizing}. We extract visual features before the last fully connected layer and Fig. \ref{fig:tsne} shows the feature distributions for source domain MNIST (red-color points) and target domain MNIST-M (blue-color points). The feature visualization aligns well with the Ablation Study in Table \ref{ablation table}. Specifically, the source and target domain features are initially not well aligned as in (a) but the alignment is improved clearly when IntraDA, InterDA, or AST is incorporated, respectively, as in (b), (c), and (d). When all three components are incorporated, we get the best feature alignment as in (e). 

In addition, to demonstrate model interpretability, we apply the Grad-CAM \cite{selvaraju2017grad} algorithm to generate class activation maps that indicate important regions in the input that lead to predictions. As illustrated in Fig. \ref{fig:cam}, by comparing the heat maps in the second row (without domain adaptation) and the third row (with CRMA), we observe that CRMA could shift the model attention to more discriminative regions within the image for the desired classification task through domain adaptation.

\section{Conclusion}
\textcolor{black}{This paper presents a concise yet effective method that exploits consistency-regularized self-training for multi-source unsupervised domain adaptation. For each source domain, we train a pair of domain-specific classifiers to perform intra-domain alignment based on the intra-domain consistency of target predictions. In addition, we compute the mean prediction of domain-specific classifier pairs to perform inter-domain alignment by maximizing the inter-domain consistency of all classifiers. As different source domains have different similarities with the target domain, we design an adaptive pseudo label generation technique that predicts target labels by weighted averaging the mean predictions of multiple source domains. Extensive experiments show that our method obtains superior accuracy consistently across all three widely studied datasets on multi-domain unsupervised domain adaptation.}

\textcolor{black}{The proposed method effectively addresses the multi-source domain adaptive classification problem with a small overhead on top of a base network model, and it could be applied to various classification or segmentation tasks when annotations are scarce or unavailable in the target domain. However, it relies on multiple classifiers and their decision boundaries for feature alignment, which limits its adaptability to more complicated tasks such as object detection. We will continue to study more generic domain adaptation techniques that can work with minimal task-specific designs in our future works.}

\bibliography{mybibfile}






\end{document}


\begin{frontmatter}

\title{Domain Consistency Regularization for Unsupervised Multi-source Domain Adaptation}

\author{Zhipeng Luo}
\address{Nanyang Technological University}
\address{Sensetime Research}

\author{Xiaobing Zhang}
\address{University of Electronic Science and Technology of China}

\author{Shijian Lu\corref{corresponding}}
\address{Nanyang Technological University}
\cortext[corresponding]{Corresponding author}

\author{Shuai Yi}
\address{Sensetime Research}




\end{frontmatter}

\linenumbers

\section*{Appendix}

In this appendix, we provide more details about the experiment setups that we do not have sufficient space to include in our submitted paper manuscript. For fair comparisons, we use the same backbone network as used in the compared methods \cite{peng2019moment, wang2020learning}. Table \ref{experiment setups} shows the training setups for the three studied datasets. In all the experiments, we use Pytorch as the deep learning framework and NVIDIA V100 GPUs for network training.

\begin{table}[h]
  \caption{The training setups for the three studied datasets: The \textit{Batch size} denotes the number of training samples in each mini-batch drawn from each domain during training.
  }
  \label{experiment setups}
  \centering
  \scalebox{0.8}{
  \begin{tabular}{c|c|c|c|c|c}
    \toprule
    Dataset & Backbone & Image size & \makecell{Batch \\ size} & \makecell{Learning \\ rate} & Epoch \\
    \midrule
    Digits-5 & Lenet & 32\times 32 & 128 & 1\times 10\textsuperscript{-3} & 50 \\
    DomainNet & Resnet-101 & 224\times 224 & 16 & 1\times 10\textsuperscript{-3} & 10 \\
    PACS & Resnet-18 & 224\times 224 & 16 & 1\times 10\textsuperscript{-3} & 100 \\
    \bottomrule
  \end{tabular}
  }
\end{table}

Specifically, Lenet \cite{lecun1998gradient} consists of three convolutional layers and three fully connected layers, and the input channels for the three fully connected layers are 8192, 3072, and 2048, respectively. For Resnet-101 and Resnet-18 \cite{he2016deep}, we use two fully connected layers following the convolution blocks with input channels of (2048, 1000) and (512, 512), respectively. In all our experiments, we treat the convolutional layers as the feature extractor $F$ and fully connected layers as the classifier $C$. The Lenet model is trained from scratch. For ResNet-18 and ResNet-101, we follow \cite{wang2020learning} and load the checkpoints that are pre-trained on ImageNet \cite{deng2009imagenet} as the feature extractor. Besides, we adopt the practice in \cite{zhu2019aligning} and assign a smaller learning rate to the pre-trained feature extractors during training, which is 1/10 of the base learning rate. To speed up the training process for the dataset DomainNet, we apply the cosine annealing scheduling \cite{loshchilov2016sgdr} to adjust the learning rate.

\bibliography{mybibfile}